\documentclass{llncs}
\usepackage{makeidx}  
\begin{document}

\mainmatter              
\title{Using the DIFF Command\\ for Natural Language Processing}
\titlerunning{NLP using DIFF}  
%
\author{Masaki Murata \and Hitoshi Isahara}

\authorrunning{Masaki Murata \and Hitoshi Isahara}   
%
\tocauthor{Masaki Murata(Communications Research Laboratory, MPT),
Hitoshi Isahara(Communications Research Laboratory, MPT)}

\institute{Communications Research Laboratory,\\ 2-2-2 Hikaridai, Seika-cho, Soraku-gun, Kyoto, 619-0289, Japan,\\
\email{\{murata,isahara\}@crl.go.jp},\\ WWW home page:
\texttt{http://www.crl.go.jp/khn/nlp/members/murata/index.html}}

\maketitle              

\def\None#1{}

\begin{abstract}
{\it Diff} is a software program that detects differences 
between two data sets and is useful in natural language processing. 
This paper shows several examples of the application of 
{\it diff}. They include the detection of differences 
between two different datasets, 
extraction of rewriting rules, 
merging of two different datasets, and 
the optimal matching of two different data sets. 
Since {\it diff} comes with any standard UNIX system, 
it is readily available and very easy to use. 
Our studies showed that 
{\it diff} is a practical tool for 
research into natural language processing. 
\end{abstract}
%

\section{Introduction}

{\it Diff}, a software program that detects differences 
between two sets of data, has numerous applications in natural language processing. 
In this paper, we first describe {\it diff} 
and then give several examples of 
how {\it diff} can be used in natural language processing. 
These examples include 
the detection of differences, 
the merging of two sets of different data, 
the extraction of rewriting rules, and 
the best matching of two different sets of data.\footnote{This 
papar is an English rough translation of our papers \cite{murata2001_mdiff_eng,murata_diff_nlp2002_eng}.}

We summarize the value of this paper below.
\begin{itemize}
\item 
  Since {\it diff} is one of the basic programs that comes with UNIX systems, 
  it is readily available and very easy to use. 
  As this paper shows, 
  {\it diff} is also applicable to various kinds of studies in the field of natural language processing. 
  This combination makes the program particularly valuable. 

\item 
  Paraphrasing has been become an active area of study \cite{murata_paraphrase}. 
  Section \ref{sec:kakikae} shows that 
  we can use {\it diff} to 
  examine the differences between 
  a spoken language and the corresponding written language 
  and in acquiring the rules for the  transformations between 
  the spoken and written languages. 
  {\it Diff} is not only useful for such studies,  
  but is also applicable to other kinds of studies in the field of paraphrasing. 
  This paper will provide a basis for such further application of {\it diff}. 

\item 
  In general, {\it diff} is used to detect differences between two data. 
  However, it can also be used to merge data 
  and to match data optimally. 
  Section \ref{sec:merge} covers several 
  interesting studies of the use of {\it diff} 
  in merging and best-matching tasks. 
  These examples are the alignment of a paper and a corresponding presentation 
  and a question-answering system. 
  This paper thus has originality 
  in terms of topics researched and 
  approach to the research. 

\end{itemize}

\section{Diff and Mdiff}
\label{sec:diff_and_mdiff}

In this section, we describe {\it diff}. 
{\it Diff} is a software program of Unix systems 
that is used to compare files. 
The program displays line-by-line differences 
between a pair of text files 
while retaining the order of the data. 
For example, suppose we have the following two files: 
\begin{verbatim}
    File 1:               File 2:
    I                     I
    go                    go
    to                    to
    school.               university.
\end{verbatim}
When we give these data to {\it diff}, 
the difference is displayed in the following way. 
\begin{verbatim}
    < school.
    > university.
\end{verbatim}

{\it Diff} has a {\it -D} option, which is very useful. 
When we use {\it diff} with this option, 
common parts as well as differences are displayed. 
That is, files can be merged by using this option. 
However, the output of {\it diff -D} is in a form 
which is used for C preprocessor such as ``Ifdef'' 
and this is difficult for people to read. 
Therefore, in this paper, we give 
the differences in the following way:
\begin{verbatim}
    ;===== begin =====
    (The parts which only exist in the first file)
    ;-----------------
    (The parts which only exist in the second file)
    ;=====  end  =====
\end{verbatim}
where, ``\verb+;===== begin =====+'' indicates 
the beginning of the differences, 
``\verb+;====+ \verb+=  end  =====+'' indicates the end of the differences, 
and ``\verb+;-----------------+'' indicates the boundary 
between the two sets of data. 
In this paper, we refer to {\it diff} in the case where
the files are merged 
by using the \verb+-D+ option and 
the differences are displayed in the above form 
as {\it mdiff}. 

When we give our earlier pair of files to {\it mdiff}, 
we obtain the following result. 
\begin{verbatim}
    I
    go 
    to 
    ;===== begin =====
    school.
    ;-----------------
    university.
    ;=====  end  =====
\end{verbatim}
``I go to'' matchs while 
``school'' and ``university'' are differences. 
The output of {\it mdiff} is 
easy to examine and understand 
because it also displays, unlike {\it diff}, the common parts. 

We can reproduce the two original files 
from the output of {\it mdiff}. 
When we take the common part and 
the upper part of the differences, 
we obtain the content of the first file. 
When we take the common part and 
the lower part of the differences, 
we obtain the content of the second file. 
We can reproduce all of the original data in this way. 

Since {\it mdiff} only displays the common part of the data once, 
it is able to reduce the amount of data. 
Since it is possible to fully reproduce the original data
from {\it mdiff}'s output, 
we are able to say that {\it mdiff} compresses the data 
while retaining the original information. 

Since the output of {\it diff} is difficult to read and 
the output of {\it mdiff} contains all information 
output by {\it diff}, 
we use {\it mdiff} for our explanations in the following sections. 
In the following sections, 
let's look at some actual examples of 
the application of {\it mdiff} to natural language processing. 

\section{Detecting differences and acquiring transformation rules}

In this section, we describe 
the studies where we used {\it mdiff} to detect differences 
and then acquired transformation rules from the differences. 
In more concrete terms, we give the following two examples. 
\begin{itemize}
\item 
  Detection of differences between the outputs of multiple systems 

\item 
  Examination of differences to 
  acquire transformation rules 

\end{itemize}

\subsection{Detection of differences between the outputs of multiple systems}

\label{sec:system}

We have been using multiple morphological analyzers 
to improve the quality of the results of morphological analysis. 

Suppose that we use two morphological analyzers 
to analyze ``We like apples.''; 
the results are as follows. 
\begin{verbatim}
    System 1                 System 2
    We       Noun            We       Noun
    like     Verb            like     Preposition
    apples   Noun            apples   Noun
\end{verbatim}
The word, ``like'' is ambiguous in terms of part of speech (POS)
and can take ``Verb'' or ``Preposition''. 
The POS of ``like'' in the above sentence is ``Verb'' 
and the analysis by System 2 was wrong.  
Here, when we give the two results to {\it mdiff}, 
we obtain the following results. 
\begin{verbatim}
    We      Noun
    ;===== begin =====
    like    Verb
    ;-----------------
    like    Preposition
    ;=====  end  =====
    apples  Noun
\end{verbatim}
{\it Mdiff} makes it easy for us 
to detect differences between the results produced by multiple systems. 
In this case, 
we notice that 
there are differences in the line for ``like''. 
Here, we determine beforehand that 
a worker corrects the outputs and judges 
which is correct when such differences are detected. 
When the output of the first system is correct
he does nothing and when the output of the second system is correct 
he marks the beginning of ``;-----------------'' with an ``x''.
When we do so, 
we can automatically select the better result of each difference 
from the results marked by the worker. 
We simply delete the lower part of the difference 
where there is no ``x'' and 
delete the upper part of the difference 
when there is an ``x''. 
As a result, 
we produce a more accurate overall result than either of the two original results. 
There are many cases when 
both parts of a difference are incorrect. 
In such cases, 
we manually rewrite the data correctly 
above ``;-----------------''. 

When we use this method, 
the only incorrect data that is not corrected 
are the data on which 
both the systems have performed exactly the same wrong analysis. 
Many errors can thus be corrected. 
What we want to say here is 
that we must prepare multiple systems that have 
different characteristics. 
If we use two systems that often produce the same wrong analysis, 
we will fail to correct a large number of errors. 

{\it Diff3}, which is another unix program, 
is available for analysing the outpus of three systems. 
{\it Diff3} detects differences among three files. 

Although we have given an example of morphological analysis, 
we can also use {\it mdiff} to detect differences 
between the results of other forms of analysis as long as 
the results are expressed in a line-by-line form. 

Although we have described a study where 
we detected differences between the outputs of multiple systems, 
we are also able to detect and correct errors in tagged corpora
by comparing the tagged corpora and the results 
of analysis of the corpora by a certain system.\footnote{There are studies on the correction of errors in tagged corpora \cite{NLP2001_eng}.}

\begin{table}[t]
  \begin{center}
    \leavevmode
    \caption{Examples of written data and 
      the corresponding data in spoken form}
    \label{tab:write_talk_juman}
\begin{tabular}{|l|}\hline
\multicolumn{1}{|c|}{The written-language sample}\\\hline
In\\
this\\
paper,\\
we\\
describe\\
the\\
meaning\\
sort.\\
In\\
general,\\
sorting\\
is\\
performed\\
by\\
using\\
\hline
\end{tabular}
\hspace{.5cm}
\begin{tabular}{|l|}\hline
\multicolumn{1}{|c|}{The spoken-language sample}\\\hline
Today\\
I'd\\
like\\
to\\
describe\\
uh\\
the\\
meaning\\
sort.\\
In\\
general,\\
sorting\\
is\\
done\\
by\\\hline
\end{tabular}
\end{center}
\end{table}

\begin{table}[t]
  \begin{center}
    \leavevmode
    \caption{The result of applying {\it mdiff} to the written-language sample 
      and the corresponding spoken-language sample}
    \label{tab:write_talk_diff}
\begin{tabular}[t]{|l|}\hline
;===== begin =====\\
In\\
this\\
paper,\\
we\\
;-----------------\\
Today\\
I'd\\
like\\
to\\
;=====  end  =====\\
describe\\
;===== begin =====\\
;-----------------\\
(Contined in the right-hand column)\\\hline
\end{tabular}
\hspace{.5cm}
\begin{tabular}[t]{|l|}\hline
uh\\
;=====  end  =====\\
the\\
meaning\\
sort.\\
In\\
general,\\
sorting\\
is\\
;===== begin =====\\
performed\\
;-----------------\\
done\\
;=====  end  =====\\
by\\\hline
\end{tabular}
\end{center}
\end{table}

\begin{table}[t]
  \begin{center}
    \leavevmode
    \caption{Extraction of differences}
    \label{tab:write_talk_diff_ext}
\begin{tabular}{|l|l|}\hline
\multicolumn{1}{|c|}{The written-language data} & \multicolumn{1}{|c|}{The spoken-language data}\\\hline
In this paper, we & Today I'd like to\\
         & uh\\
performed & done\\\hline
\end{tabular}
\end{center}
\end{table}

\subsection{Examining differences and acquiring transformation rules} 
\label{sec:diff_spoke_written}
\label{sec:kakikae}

In this section, we describe our study of the use of 
{\it mdiff} to detect differences 
between a spoken language and the corresponding written language. 
In this study, 
we used the alignment between data on  
the spoken language and on the written language 
to examine the differences between 
the spoken and written languages 
on the basis of the results of {\it mdiff}. 
We thus acquired rules 
for transforming the spoken language to the written language 
along with the inverse  transformation rules. 
We used presentations at academic conferences 
as examples of the spoken language 
and used papers which had the same content as the presentations 
as examples of the writtten language. 

For example, we supposed that 
data in the spoken and written languages 
are given as in Table \ref{tab:write_talk_juman}.\footnote{Although we have given English sentences in the table, 
we actually performed these experiments on Japanese-language materials.}
The data in the table are transformed 
so that each line has one word 
before we apply {\it mdiff}. 
Applying {\it mdiff} to the data 
produces the results shown in Table \ref{tab:write_talk_diff}. 
By extracting the differences from these results, 
we obtain the results shown in Table \ref{tab:write_talk_diff_ext}. 

The results show us that 
``uh'' is inserted in the spoken language 
and that ``performed'' can, in this case, be paraphrased 
as ``done''.
We can use {\it mdiff} to
detect differences between 
the spoken language and the written language and then 
examine them linguistically. 
We can also consider the detected differences 
as rules for the transformation between written 
and spoken languages. 
For example, the occurrence of ``uh'' can be 
regarded as indicating a rule that 
we may insert ``uh'' into the written-language data 
as part of transforming it into the spoken language. 
The difference in terms of whether ``performed'' or ``done'' 
is used can be regarded as a rule that 
we transform ``performed'' to ``done'' 
when we transform data from the written 
to the spoken language. 
We thus find that 
we are able to use {\it mdiff} to detect the transformation rules. 

In this section, 
we used data in the written and spoken language. 
However, we can study a variety of data in a similar way. 
For example, 
when we apply {\it mdiff} 
to written texts before and after spelling and
grammar are checked, 
we can find out how we should modify the sentences to eliminate spelling 
and grammatical errors and 
thus acquire rules for checking spelling and grammar. 
When we use {\it mdiff} 
on texts and the results of their automatic summarization, 
we can clearly see how the texts were summarized 
and can thus acquire the rules used in automatic summarization. 
In other cases of paired textual data, too, 
we will be able to apply {\it mdiff} to examine 
the data in the various ways that such pairs can be obtained and 
to obtain the rules for transformation. 

\section{The merging and best matching of two different sets of data}
\label{sec:merge}

In this section, we describe 
studies of the use of {\it diff}'s function 
of optimally merging data.\footnote{In {\it diff}, 
best matching is performed by maximizing 
the extent of the common parts.}
We describe the following two examples. 
\begin{itemize}
\item 
  The alignment of presentations and corresponding papers
\item 
  A question-answering system using 
  the function of finding the the best match
\end{itemize}

\begin{figure}[t]
  \begin{center}
    \leavevmode

\begin{minipage}[h]{4cm}
\verb+<Chapter 1>+

(the contents of Chapter 1)

\verb+</Chapter 1>+

\verb+<Chapter 2>+

(the contents of Chapter 2)

\verb+</Chapter 2>+

\verb+<Chapter 3>+

(the contents of Chapter 3)

\verb+</Chapter 3>+
\end{minipage}

\caption{Structure in the paper}
\label{tab:youkou_kousei}
\end{center}
\end{figure}

\begin{figure}[t]
  \begin{center}
    \leavevmode
    \begin{tabular}[t]{|l|}\hline
;===== begin =====\\
\verb+<Chapter 1>+\\
(content of the paper only)\\
;-----------------\\
(content of the presentation only)\\
;=====  end  =====\\
(content of the paper and presentation)\\
;===== begin =====\\
(content of the paper only)\\
;-----------------\\
(Contined in the right-hand column)\\\hline
    \end{tabular}
\hspace{0.5cm}
    \begin{tabular}[t]{|l|}\hline
(content of the presentation only)\\
;=====  end  =====\\
(content of the paper and presentation)\\
;===== begin =====\\
\verb+</Chapter 1>+\\
\verb+<Chapter 2>+\\
(content of the paper only)\\
;-----------------\\
(content of the presentation only)\\
;=====  end  =====\\\hline
\end{tabular}

\caption{Results of applying {\it mdiff} to a paper and a corresponding presentation}
\label{tab:youkou_mdiff}
\end{center}
\end{figure}

\begin{figure}[t]
  \begin{center}
    \leavevmode

\begin{minipage}[h]{10cm}
\verb+<Chapter 1>+

(content of the presentation only)

(content of the paper and presentation)

(content of the presentation only)

\verb+</Chapter 1>+

\verb+<Chapter 2>+

(content of the presentation only)

\end{minipage}

\caption{Results of insertion of information on sections in a presentation}
\label{tab:youkou_mdiff2}
\end{center}
\end{figure}

\subsection{Alignment of a paper with the corresponding presentation}
\label{sec:sp_merge}

In this section, we align
a paper with the corresponding presentation \cite{uchimoto2001_eng}. 
The paper and corresponding presentation is the same example 
as was used in describing the acquisition of 
the rules for transformation from written to spoken-language data 
(Section \ref{sec:diff_spoke_written}).
The presentation was made at an academic conference and 
the paper corresponds to this presentation. 
When we are able to align each part of a presentation 
with a corresponding part of the paper, 
this form of alignment is very useful. 
For example, 
when we listen to a presentation, 
we are able to refer to the part of the paper that corresponds 
to the part of we are now listening. 
When we read a paper, 
we are able to refer to the part of the presentation that corresponds 
to the part of the paper we are now reading \cite{uchimoto2001_eng}. 
In this section, 
we consider the use of {\it mdiff} to align a paper with the corresponding presentation. 

Here, let's try to determine 
the parts of the presentation to which each chapter 
of the paper corresponds by using {\it mdiff}. 
We suppose that 
the content is laid out in the same order in the paper and in the presentation. 
In advance, we place 
symbols such as \verb+<Chapter 1>+, as shown in Figure \ref{tab:youkou_kousei}, 
into the paper. This is so that we are easily able to recognize 
the chapters of the paper. 
By applying {\it mdiff} to the data set that 
consists of a presentation after both have been 
transformed so that each line has one word, 
we obtained the results shown in Figure \ref{tab:youkou_mdiff}. 
Next, we obtained the results shown in Figure \ref{tab:youkou_mdiff2} 
by eliminating the upper parts of the differences, i.e., those that correspond
to the paper, and 
leaving symbols such as \verb+<Chapter 1>+ in place. 
The symbols such as \verb+<Chapter 1>+ 
are only inserted in the data of the presentation. 
We are then able to recognize 
the chapter of the paper that corresponds with a given part 
of the presentation. 

To put this simply, 
we place information to indicate chapters 
in the data of presentations. 
We then use the merging function of {\it mdiff} 
to match the parts of the paper and presentation. 
Then, by eliminating the content of the paper, 
we are left with information on the chapters. 
We can easily align a paper with a corresponding presentation 
at the level of chapters by using {\it mdiff}. 

\subsection{A question-answering system using 
best matching}

In this section, we describe 
our question-answering system 
that utilizes {\it mdiff}'s best-matching function \cite{murata_paraphrase,qa_memo}. 

A question-answering system receives a question 
and outputs an answer. 
For example, when ``Where is the capital of Japan?'' 
is given, it returns ``Tokyo''. 
When we consider 
now much knowledge is in written natural language form, 
we see that the system only has to match a given question 
with a sentence (a knowledge sentence) that includes the knowledge then select the word 
in the sentence that corresponds to the 
interrogative pronoun of the question sentence. 
For example, 
in the case of the previous question, 
the system detects this knowledge sentence
in a knowledge database:  ``Tokyo is the capital of Japan.'' 
``Tokyo'' is output as the answer 
since corresponds to the interrogative pronoun in the sentence. 
Here, we consider the use of {\it mdiff} in answering questions. 

\begin{figure}[t]
  \begin{center}
    \leavevmode

\begin{minipage}[t]{4cm}
\begin{verbatim}
    ;===== begin =====
    X
    ;-----------------
    Tokyo
    ;=====  end  =====
    is
    the 
    capital
    of 
    Japan. 
\end{verbatim}
\vspace{1.5cm}
\hspace{1cm}
(a) Case 1
\end{minipage}
\hspace{1cm}
\begin{minipage}[t]{4cm}
\begin{verbatim}
    ;===== begin =====
    X
    ;-----------------
    Tokyo
    ;=====  end  =====
    is
    the 
    capital
    ;===== begin =====
    of 
    ;-----------------
    in
    ;=====  end  =====
    Japan. 
\end{verbatim}
\hspace{1cm}
(b) Case 2
\end{minipage}

\caption{Results of {\it mdiff} in the question-answering system}
\label{tab:qa_result}
\end{center}
\end{figure}

Firstly, we transform 
the interrogative pronoun in the question sentence into X 
and change ``?'' into ``.'' at the end of the sentence. 
We then obtain ``X is the capital of Japan.''. 
We thus obtain ``Tokyo is the capital of Japan.'' from the knowledge base. 
We obtain the results in Figure \ref{tab:qa_result} (a) 
by applying {\it mdiff} to the two sentences. 
We consider the part that is paired with X in the differences 
as the answer and correctly obtain ``Tokyo''. 

Even if there are differences between the compared sentences, 
we can still use {\it mdiff} to correctly extract the answer. 
For example, suppose that 
the knowledge sentence is ``Tokyo is the capital in Japan.'' 
In this case, the result of {\it mdiff} is as in Figure \ref{tab:qa_result} (b). 
Although the number of differences increased, 
the part that corresponds to X remains ``Tokyo'' and 
the correct answer has been detected. 

The question-answering system we have proposed 
repeats to transform the question sentence and 
knowledge sentence so that 
the two sentences are as similar as possible. 
It then extracts the answer by matching 
the sentences when they are at this point. 
Since we use the similarity between two sentences in that time, 
we have to define the degree of the similarity between sentences.
Since we can recognize the common parts and 
the different parts by using {\it mdiff}, 
we can calculate the degree of similarity 
as (the number of characters in the common part)/(the number of 
all characters).\footnote{Here, 
we use {\it mdiff} to calculate the degree of similarity of the sentences. 
{\it Mdiff} is also useful in this way.}
Here, if we suppose that 
we have a rule for transforming ``in'' to ``of'', 
we match the data 
after transforming ``Tokyo is the capital in Japan.'' 
to ``Tokyo is the capital of Japan.'' 
and we are able to obtain the answer more reliably 
by decreasing the number of differences. 

\section{Conclusion}

In this paper, we describe a lot of examples of 
the use of {\it diff} in various problems of 
natural language processing. 
In Section \ref{sec:system}, 
we describe how we apply {\it diff} 
to a system that combines multiple systems 
and thus obtain 
higher precisions than from any of individual systems. 
Studies on combination of systems are very well known. 
We describe how such combination can easily be performed by {\it diff}. 
In Section \ref{sec:kakikae}, we showed that 
{\it diff} is easily able to examine the differences 
between a spoken and written language 
and acquire rules for the transformation 
between the spoken and written language. 
We handled the paraphrasing of 
written language into spoken language. 
However, the study on paraphrasing 
covers a wide range: 
automatic extraction of compressed sentences, 
sentence polishing-up to modify sentences correctly, 
and transformations 
from difficult sentences to plain sentences. 
In these studies, too, 
it will be easy to apply {\it diff} to 
various forms of examination and to the extraction of various transformation rules. 
This paper will become a basis for studies of paraphrasing, 
an area in which there has been increasing activity. 
In Section \ref{sec:merge}, 
we showed examples of the merging of data and 
the optimal matching of data. 
Since {\it diff} is generally used to detect differences, 
Section \ref{sec:merge}, which shows 
applications of {\it diff} to merging and matching, 
presents a new and original approach. 
In the section, we showed that 
it is easy to apply {\it diff} to 
interesting studies of two kinds, 
the alignment of a paper with a corresponding presentation 
and a question-answering system. 

In this paper, we have described many examples 
of the application of {\it diff} to natural language processing. 
We hope that {\it diff} will be applied to an ever-widening range of studies. 

\section*{Acknowlegement}

We were given 
data for the experiments on alignment 
of a paper with a corresponding presentation 
by Kiyotaka Uchimoto of the Communications Research Laboratory.


\begin{thebibliography}{1}

\bibitem{murata2001_mdiff_eng}
Masaki Murata and Hitoshi Isahara.
\newblock Nlp using diff.
\newblock {\em IPSJ-WGNL 2001-NL-144}, 2001.
\newblock (in Japanese).

\bibitem{murata_paraphrase}
Masaki Murata and Hitoshi Isahara.
\newblock Universal model for paraphrasing: Using transformation based on a
  defined criteria.
\newblock In {\em NLPRS'2001 Workshop on Automatic Paraphrasing: Theories and
  Applications}, 2001.

\bibitem{murata_diff_nlp2002_eng}
Masaki Murata and Hitoshi Isahara.
\newblock {NLP} using {DIFF} --- use of convenient tool for detecting
  differences, {MDIFF} ---.
\newblock {\em Journal of Natural Language Processing}, 9(2), 2002.
\newblock (in Japanese).

\bibitem{qa_memo}
Masaki Murata, Masao Utiyama, and Hitoshi Isahara.
\newblock Question answering system using syntactic information.
\newblock 1999.
\newblock http://xxx.lanl.gov/abs/ cs.CL/9911006.

\bibitem{NLP2001_eng}
Masaki Murata, Masao Utiyama, Kiyotaka Uchimoto, Qing Ma, and Hitoshi Isahara.
\newblock Correction of the modality corpus for machine translation based on
  machine-learning method.
\newblock {\em 7th Annual Meeting of the Association for Natural Language
  Processing}, 2001.
\newblock (in Japanese; an English translation of this paper is available at
  http://arXiv.org/abs/cs/0105001).

\bibitem{uchimoto2001_eng}
Kiyotaka Uchimoto, Chikashi Nobata, Kimiko Ohta, Masaki Murata, Qing Ma, and
  Hitoshi Isahara.
\newblock Segmenting the transcription of a talk by aligning the transcription
  with its corresponding paper.
\newblock {\em 7th Annual Meeting of the Association for Natural Language
  Processing}, pages 317--321, 2001.

\end{thebibliography}
\end{document}